\documentclass[conference]{IEEEtran}
\IEEEoverridecommandlockouts
\usepackage{cite}
\usepackage{amsmath,amssymb,amsfonts}
\usepackage{algorithmic}
\usepackage{graphicx}
\usepackage{textcomp}
\usepackage{xcolor}
\def\BibTeX{{\rm B\kern-.05em{\sc i\kern-.025em b}\kern-.08em
    T\kern-.1667em\lower.7ex\hbox{E}\kern-.125emX}}
\begin{document}

\title{Evaluating the Rainbow DQN Agent in Hanabi with Unseen Partners}


\author{\IEEEauthorblockN{Rodrigo Canaan}
\IEEEauthorblockA{
\textit{New York University}\\
New York, USA \\
rodrigo.canaan@nyu.edu}
\\
\IEEEauthorblockN{Julian Togelius}
\IEEEauthorblockA{
\textit{New York University}\\
New York, USA \\
julian.togelius@nyu.edu}
\\
\and
\IEEEauthorblockN{Xianbo Gao}
\IEEEauthorblockA{
\textit{New York University}\\
New York, USA \\
xg656@nyu.edu}
\\
\IEEEauthorblockN{Andy Nealen}
\IEEEauthorblockA{
\textit{University of Southern California}\\
Los Angeles, USA \\
anealen@cinema.usc.edu}
\and
\IEEEauthorblockN{Youjin Chung}
\IEEEauthorblockA{
\textit{New York University}\\
New York, USA \\
yjc433@nyu.edu}
\\
\IEEEauthorblockN{Stefan Menzel}
\IEEEauthorblockA{
\textit{HRI Europe GmbH}\\
Offenbach, Germany \\
stefan.menzel@honda-ri.de}
}


\maketitle

\begin{abstract}
Hanabi is a cooperative game that challenges existing AI techniques due to its focus on modeling the mental states of other players to interpret and predict their behavior. While there are agents that can achieve near-perfect scores in the game by agreeing on some shared strategy, comparatively little progress has been made in ad-hoc cooperation settings, where partners and strategies are not known in advance. In this paper, we show that agents trained through self-play using the popular Rainbow DQN architecture fail to cooperate well with simple rule-based agents that were not seen during training and, conversely, when these agents are trained to play with any individual rule-based agent, or even a mix of these agents, they fail to achieve good self-play scores. 
\end{abstract}





\section{INTRODUCTION}

\noindent Cooperative multi-agent problems with hidden information are challenging for humans and AI systems due to the need to model other actors' mental states. This model can be used both to predict their future behavior and to infer unseen features of the world through the lens of their observed behavior. The ability to impute distinct mental states to oneself and others has been referred to as having a theory of mind~\cite{premack1978does}. Hanabi (Antoine Bauza, 2010) is a cooperative card game that has received attention of AI researchers because strategies for playing it rely heavily on theory of mind and communication.

While agents that achieve near-perfect scores in a self-play setting using a shared strategy have been developed for the game~\cite{bouzy2017playing,foerster2018bayesian,Wu}, comparatively little progress has been made on \textit{ad-hoc cooperation} settings, where the behavior of other agents is not known in advance. In particular, there are to our knowledge no Reinforcement Learning (RL) agents designed to play either with humans or with simple rule-based agents inspired by human play such as the ones described by Walton Rivers \textit{et al.}~\cite{walton2017evaluating} and used as part of the evaluation of the CoG Hanabi competition~\cite{walton2019competition}.

In this paper, we examine the behavior of RL agents trained using the Rainbow DQN architecture~\cite{hessel2018rainbow} when paired with the aforementioned rule-based agents. In order to achieve that, we re-implemented the agents from~\cite{walton2017evaluating} into the Hanabi Learning Environment (HLE)~\cite{bard2020hanabi}, where the Rainbow DQN agent was first displayable. By making our implementation public, we hope to facilitate future comparisons between results of researchers using the CoG competition framework and the HLE.

The Rainbow DQN agent was chosen because other successful RL agents such as the Bayesian Action Decoder ($BAD$)~\cite{foerster2018bayesian} and the Actor-Critic Hanabi Agent ($ACHA$)~\cite{bard2020hanabi} have been noted to achieve high scores in self-play, but perform poorly in the Ad-Hoc scenario, and even when paired with independent instances of agents trained with the same procedures. Different instances of these agents were observed to learn policies relying on different arbitrary conventions (such as using color hints to indicate that a card in a certain slot is playable).

With the Rainbow agent, however, the authors noted that different instances often learned similar policies and played well with each other, but without quantifying these results. With this in mind, it would be possible that the Rainbow agents were learning policies that don't depend heavily on any particular convention and thus might play well with other agents such as rule-based agents following simple heuristics.

The main question we address then is: can the Rainbow DQN agent from~\cite{bard2020hanabi}  cooperate well with partners that were not seen during training? We answer this question negatively in two ways: first, we show that Rainbow agents trained purely through self-play perform very poorly when paired with the rule-based agents we selected from~\cite{walton2017evaluating}.

Second, we show that Rainbow agents that were trained with one or more rule-based agents as partners fail to play well with a particular ``unseen'' partner: itself. In other words, it fails to perform well in self-play, despite being able to achieve reasonable scores with its training partners.

This shows that the Rainbow DQN agent, despite learning policies that work well in self-play and across independently-trained instances, is unable to perform well with other agents it has not seen during training.

 
\section{The Hanabi Card Game}

Hanabi is a cooperative game for two to five players, who draw from a deck of cards featuring a color and a numerical rank. The goal is to collectively play cards from each color in ascending order of rank, making one pile for each color.  However, each player can only see cards in the other players' hands, but not their own cards. If a card is played in the wrong order, the group loses a life, and if three lives are lost the game is over. 

Communication between players is not allowed except for a limited number of hint actions. Each hint action costs an information token from a shared pool and allows the active player to choose one of the other players and point to all the cards with a chosen color or rank on that player's hand. A player may also discard a card from their hand in order to recoup an information token for the group, but in doing so risks discarding a card that was necessary to complete one of the piles (as the number of cards with each rank and color in the deck is limited). When the deck runs out, players take one last turn each, and then the game is over. 

The group scores one point for each card played correctly, up to a maximum of 25, corresponding to five cards played for each of the five color. If the group loses all lives it scores zero regardless of its partial score up to that point. We call this the ``strict'' scoring scheme. Some previous research, such as~\cite{walton2017evaluating}, use an alternative ``lenient'' scoring scheme where the group keeps their partial score even if all lives run out. Unless otherwise noted, all scores reported in this paper correspond to the strict scheme and the 2-player version of the game.

Because hint actions are limited, players are encouraged to convey as much actionable information as possible with each hint. This makes playing Hanabi with humans an exercise in Theory of Mind, as players are constantly trying to imagine the world from other players' perspective and infer the intents behind each action. This sometimes leads to the development of implicit or explicit conventions (an assumption of how a player's private information affects their behavior) that help interpret actions from other players and predict their future behavior. And example of a simple convention is that, all else being equal, players should give hints relating to cards that are immediately playable. This type of convention can emerge between humans even without explicit agreement and enable players successfully play cards even if its rank or color is missing, saving hint tokens in the process.

However, with explicit coordination it is also possible to implement more complex conventions, such as using hints of some arbitrary color to indicate that the card in some arbitrary slot in a player's hand is playable. These can lead to very strong performance by a coordinated group, but can easily backfire if members have no prior coordination and don't know the exact convention being used. The fact that different conventions are not necessarily compatible makes playing Hanabi in an Ad-Hoc setting much more difficult, both for humans and for AI, than playing in a setting with prior coordination, such as learning through self-play.

\section{Related Work}

\subsection{Hanabi-Playing Agents}

The first agents for playing Hanabi are by Osawa~\cite{osawa2015solving}, whose best agent attempts to play cards that are known to be playable (``safe''), discard cards that are known to never be playable again (``useless'') and give hints about playable cards owned by other players, keeping track of the information other players already know to avoid redundant hints.

Since then, many researchers have addressed AI problems using Hanabi as testbed. Some of the most successful~\cite{cox2015make,bouzy2017playing} are based on a ``hat-guessing'' convention which, by agreeing on a clever encoding of the meaning of hints that can be interpreted by the whole table at once (not just the receiving player), achieve average scores above 24 on the 4- and 5-player versions of the game. 

Eger and colleagues created an ``intentional'' agent for Hanabi, which is meant to be easily interpretable, and evaluated it with humans~\cite{eger2017intentional}. Other work such as ~\cite{gottwald2018see} and ~\cite{eger2019wait} explore the use of external communication channels, such as eye gaze or the timing between actions which could (wittingly or unwittingly) play a role in games involving humans.

While most of these authors developed (and in some cases, made publicly available) their own implementations of the game, two environments have emerged and achieved wide use among researchers: the CoG (previously CIG) competition Java environment~\cite{walton2019competition}, which combines many of the previous agents in a rule-based paradigm and the Hanabi Learning Environment (HLE)~\cite{bard2020hanabi}, focused on reinforcement learning.

\subsection{Rule-Based Agents and the CoG competition environment}

The CoG Hanabi competition~\cite{walton2019competition} provides an environment, written in Java, for simulating the game, which also allows the user to specify a new agent by simply providing an ordered list of rules which the agent should follow. Their environment comes with enough rules to reproduce various agents in the literature up to that point, plus some agents implemented by the authors~\cite{walton2017evaluating}. Users of the environment can also specify new rules. Out of the agents considered in~\cite{walton2017evaluating}, the best at self-play is $Piers$, which (in our Java reimplementation) achieves a score of 17.31. They  organized a competition, which ran in 2018 and 2019 at the CIG/CoG conferences using this environment, which we will refer to from now on as the Java environment.

Building upon the rule-based paradigm, a method for creating new agents using a Genetic Algorithm to search for a sequence of rules that maximizes score either in the self-play setting or when paired with other rule-based agents was proposed by Canaan \textit{et al.}~\cite{canaan2018evolving}. The method was expanded in ~\cite{canaan2019diverse} to procedurally generate agents which display behaviorally diverse behaviors using MAP-Elites~\cite{mouret2015illuminating}. Agents generated through such a process could be useful for the purposes of training RL agents, but, for this paper, would require us to re-implement a larger number of rules from the Java environment into the Python environment than the ones needed to reproduce the seven rule-based agents discussed. Another work that builds upon the Java environment is the Monte Carlo Tree Search (MCTS) agent by Goodman~\cite{goodman2019re}, winner of the 2018 and 2019 competitions. This agent achieves a score of 20.5 at self-play in the lenient scoring scheme.

\subsection{The Hanabi Learning Environment}

Bard and colleagues recently introduced an environment for Reinforcement Learning in Hanabi written in Python~\cite{bard2020hanabi}, which we will refer to as the Python environment. The paper introduces a Rainbow DQN agent ($Rainbow$) and an Actor Critic Hanabi Agent ($ACHA$). $Rainbow$ is based on an architecture by Hessel \textit{et al.}~\cite{hessel2018rainbow} which achieved state-of-the-art perfomance on the Arcade Learning Environment~\cite{bellemare2013arcade} by combining multiple improvements to the DQN variant of Q learning~\cite{mnih2013playing}. In Hanabi, Rainbow achieves a self-play score of 20.64 after 100 million training steps. 

$ACHA$ is based on the Importance Weighted Actor-Learner architecture~\cite{espeholt2018impala} and achieves a self-play score of 22.73 in Hanabi with 10 billion training steps.

At the time the experiments described in this paper were run, the state-of-the-art for 2-player Hanabi was the Bayesian Action Decoder ($BAD$), described by Foerster~\cite{foerster2018bayesian}. It achieves a self-play score of 24.17 with 16.3 billion training steps by exploring the space of deterministic policies.

While $ACHA$ and $BAD$ are more powerful than $Rainbow$, we chose to use $Rainbow$ as basis of this work for three reasons: first, it is the only one with an open-source implementation (which accompanies the Python environment). Second, it is shown to achieve good results with a much more modest training budget. Finally, it is the only one of the three that we deemed to have any possibility of cooperating well with the rule-based agents we're interested in; both $BAD$ and $ACHA$ have been observed to make use of exotic conventions (such as using a color hint to indicate that an arbitrary card is playable or should be discarded). $BAD$ is designed from the ground up to learn a shared convention, and $ACHA$ was observed to learn different conventions with each training run, and as a result different $ACHA$ agents play very badly with each other.

More recently, a simplified version of $BAD$ called the Simplified Action Decoder ($SAD$) was proposed by Hu and Foerster~\cite{hu2019simplified}, which, augmented with the multi-agent search procedure described by Lerer \textit{et al.}~\cite{lerer2019improving}, achieves the new state-of-the art self-play average score of 24.6. Another variation of $SAD$ called ``Other Play'' takes into account a set of ``simmetries'' provided by humans (such as a permutation of color labels)~\cite{hu2020other}. It is able to perform well across independently trained and was shown to perform better with humans than baseline $SAD$. However, while this variation avoids learning conventions such as ``a hint of yellow indicates that the fifth card should be playable'', it still sometimes learned conventions such as ``any color hint (as opposed to a rank hint) indicates that the fifth card should be played''. While this is a promising approach to the problem of Ad-Hoc cooperation, due to the recency of agents based on $SAD$, they were not considered as candidates for our experiments.

\section{Experiments}

All experiments are based on the two-player version of the game, and we used the strict scoring scheme (where the group scores zero if it runs out of lives) during both training and evaluation.

Experiments were run on a variety of computers: a macOS Mojave iMac with a 3.8GHz Intel Core i5 processor (quad-core), a macOS Sierra MacBook Pro with a 2.9 Intel Core i7 processor (quad-core), a Linux machine with a 3.5 GHz Intel Core i7-6950X Extreme Edition (10 core) and Cuda 5.1 with three Titan X GPUs and a Linux machine with a 3.5 GHz Intel Core i7-5930K 3.5 GHz and Cuda 5.1 with three GTX 1080 GPUs . Performance varied significantly depending on the computer being used, the experiment being considered and how many processes (including some not related to this paper) were run at a time on each computer, with agents training at anywhere from 100 to 400 steps per second.

The Java and Python code for all experiments is available at https://github.com/rocanaan/hanabi-ad-hoc-learning


\begin{table}
\centering
\begin{tabular}{ccc}
\textbf{Rule-based agent} & \textbf{Self-Play (Java)} & \textbf{Score (Python)} \\
$IGGI$ & 15.77 & 15.76 \\
$Internal$ & 11.17 & 10.01 \\
$Outer$ & 14.56 & 13.78 \\
$LegalRandom$ & 0.00 & 0.00 \\
$VDB$ & 13.39 & 16.12 \\
$Flawed$ & 0.00 & 0.00 \\
$Piers$ & 17.31 & 17.06 \\
\end{tabular}

\caption{Comparison of scores between the Java and Python implementations of the seven rule-based agents. The largest Standard Deviation (SD) of any score is 4.9 and the largest Standard Error of the Mean (SEM) is 0.16.}
\label{table:validation}
\end{table}
\subsection{Re-implementation of Rule-based Agents}\label{subsection:reimplementation}


\begin{table*}
\begin{center}
\begin{tabular}{ccccccccc}
& $IGGI$ & $Internal$ & $Outer$ & $LegalRandom$ & $VDB$ & $Flawed$ & $Piers$ & Average\\
$IGGI$ & 15.87 & 12.48 & 15.25 & 0.00 & 16.50 & 0.15 & 16.85 & 11.20\\
$Internal$ & 12.48 & 10.20 & 11.81 & 0.00 & 13.39 & 0.01 & 13.67 & 8.79 \\
$Outer$ & 15.25 & 11.81 & 13.79 & 0.00 & 14.85 & 0.04 & 15.65 & 10.20 \\
$LegalRandom$ & 0.01 & 0.00 & 0.00 & 0.00 & 0.01 & 0.00 & 0.02 & 0.01 \\
$VDB$ & 16.50 & 13.39 & 14.85 & 0.01 & 16.06 & 0.15 & 17.23 & 11.17 \\
$Flawed$ & 0.15 & 0.01 & 0.04 & 0.00 & 0.15 & 0.00 & 0.16 & 0.07 \\
$Piers$ & 16.85 & 13.67 & 15.65 & 0.02 & 17.23 & 0.16 & 16.92 & 11.50\\
\end{tabular}
\end{center}
\caption{Results obtained when pairing rule-based agent with one another. In the diagonal, the self-play performance for each of the agents. The maximum SD and SEM are 4.77 and 0.15 respectively.}
\label{table:rulebased matchups}
\end{table*}

Our first step was to re-implement, in the Python environment, the seven rule-based agents Walton-Rivers and coauthors use as baseline for ad-hoc play in~\cite{walton2017evaluating}. Some of these agents were implemented for that paper, while others were themselves re-implementations of agents that were previously published by other authors. We give a brief overview of these agents' behavior below:

$LegalRandom$ is the simplest of the rule-based agents. It simply picks one of the legal actions at random to play at each turn.

$Internal$ and $Outer$ were originally designed by Osawa~\cite{osawa2015solving} and both prioritize playing a card known to be safe, followed by discarding a card known to be useless, followed by giving a hint (prioritizing playable cards), followed by discarding randomly. The difference is that $Internal$ does not keep track of other player's knowledge about their cards, and is therefore liable to giving repeated hints.  $Outer$ keeps track of this knowledge and will always provide new information with each hint.

${VDB}$ was originally designed by van den Bergh~\cite{van2016aspects} by using game simulations to explore variations in the high-level strategy taken by the Osawa agents. Among these variations, the agent will take risks in playing cards that are not guaranteed to be safe, as long as the probability of being playable is greater than a certain threshold (empirically determined to be 60\%). The agent will also give hints about useless cards or will give hints about as many cards as possible if no playable card can be hinted at.

$IGGI$ and $Piers$ were first introduced in~\cite{walton2017evaluating}. $IGGI$ only plays safe cards, and also prefers to discard its oldest card (the one that has been held for the longest time in hand) if no card in its hand is known to be useful. This potentially makes it more predictable to its partners. $Piers$ uses the same playability threshold of  as $VDB$ (60\%), but will also play its most likely playable card at the last round of the game if more than one life is left (in an attempt to score one extra point) and will only give hints about useless cards if there are fewer than 4 information tokens left.

Finally, $Flawed$ was also introduced in~\cite{walton2017evaluating} to be deliberately bad when playing with agents who aren't adapting to its behavior. It achieves this by giving hints at random and also by playing its most likely playable card with a threshold of just 25\%. It is intended to play reasonably well with agents that give it lots of information, but will tend to play in a very risky manner otherwise.

Although all of these agents can be described at a high level in a few sentences, there is surprising nuance in their implementation. For example, some rules for playing a safe card differ in whether they consider only ``positive'' information from received hints or also ``negative'' information from the same hints (cards identified as not being a certain color or rank) or information available by a process of elimination after counting the cards visible in the discard pile and other players' hand. Many hint rules also differ in how they handle cases where multiple hints would satisfy the rule: some break the tie randomly, some consider what information is already known by the player and some prefer hinting about colors than ranks (or vice-versa). We attempted to be as faithful as possible to the rules as implemented in the Java environment, but it is possible that some nuances might have been missed, leading to small discrepancies.

For the remainder of the paper, we will use the agents' name with a $J$ or $P$ subscript to denote either the original Java implementation or our re-implementation of a rule-based agent e.g. $Piers_J$ (Java) or $Piers_P$ (Python).

\subsection{Self-Play Rainbow Agents}


\begin{table*}
\begin{center}
\begin{tabular}{ccccccc}
& $Rainbow_{SP1}$ & $Rainbow_{SP2}$ & $Rainbow_{SP3}$ & $Rainbow_{SP4}$ & $Rainbow_{SP5}$ & Average \\
$Rainbow_{SP1}$ & 17.79 & 19.01 & 18.30 & 19.11 & 18.93 & 18.63 \\
$Rainbow_{SP2}$ & 19.01 & 18.95 & 18.31 & 19.22 & 18.49 & 18.80\\
$Rainbow_{SP3}$ & 18.30 & 18.31 & 18.13 & 19.33 & 18.74 & 18.56\\
$Rainbow_{SP4}$ & 19.11 & 19.22 & 19.33 & 19.19 & 18.53 & 19.08\\
$Rainbow_{SP5}$ & 18.93 & 18.49 & 18.74 & 18.53 & 18.70 & 18.68\\
\end{tabular}
\end{center}
\caption{Results obtained when pairing each version of $Rainbow_{SP}$ agents with one another. In the diagonal, the self-play performance
for each of the agents. The maximum SD and SEM are 4.2 and 0.13 respectively.}
\label{table:SelfPlay}
\end{table*}

We  trained five independent instances of $Rainbow$ in the self-play regime. We denote those by $Rainbow_{SP_i}$ where $i$ is an index from 1 to 5. Although Bard \textit{et al.}~\cite{bard2020hanabi} have previously trained Rainbow agents using the exact same procedure, they were either validated in self-play mode or with instances of $ACHA$ as partners, where they were shown to not cooperate well. As previously discussed, they were shown to achieve very low scores when paired with $ACHA$, but it is unclear if this was simply because $ACHA$ is a bad cooperator or if $Rainbow$ itself is also not good at playing with other strategies. The authors have noted, however, that $Rainbow$ seems to converge on the same strategy every time it is trained, but did not provide numerical results showing how well independently trained instances of $Rainbow$ play together.

The architecture we use for this paper is the same as used in~\cite{bard2020hanabi}: it is a 2-layer Multi-Layer Perceptron with 512 nodes per hidden layer, that takes a vectorized  representation of the game state composed of 658 binary values and outputs one of the up to 20 legal actions to take.  All hyperparameters were the same as those used in the original paper.

The game state representation used encodes features of the game that are directly seen by the agent, such as the cards already played and discarded, the number of hint tokens, life tokens and cards in the deck, and the rank and color of all cards in other player's hand. The representation also keeps track of hints received by all players, so the agent knows at all times the possible rank and color of cards in its hand and also which information is known or missing from other players.

However, the representation doesn't keep track of game history beyond the last action, and the neural network used by the agent is feed-forward with no recurrent connections. As such, the agent is not expected to learn policies dependent on a long history of actions. As such, our inquiry on how well this agent plays with other teammates is not out of hope that it is able to identify and adapt to different teammates, since it lacks the long-term memory needed for it. Rather, it is based on the observation that the agent seems to learn strategies that work well among independently trained instances of the policy, which might be a sign that these strategies rely less on arbitrary conventions and more on grounded information, which might lead them to play well with the rule-based agents we chose, which also mostly use only grounded information.

So our training of $Rainbow_{SP}$  serves two purposes: first, it enables us to establish numerically  whether $Rainbow$ learns strategies that play well together from run to run, and secondly and most importantly, whether it plays well with human-inspired strategies such as those used by the rule-based agents. To our knowledge, the only other work where a learning agent (as opposed to rule-based or those based on tree-search) has been paired with human-inspired agents for evaluation has been in~\cite{bard2020hanabi},where $ACHA$ was paired with SmartBot~\cite{O'Dwyer}, achieving a score of effectively zero.

$Rainbow_{SP1}$ was trained for 37.5 million steps. $Rainbow_{SP2}$ was trained for 55 million steps. $Rainbow_{SP3}$, $Rainbow_{SP4}$ and $Rainbow_{SP5}$ were trained for 47.5 million steps each.

\subsection{Agents Paired with Known Partners for Training}

Our next step was to train RL agents that are specialized at playing with each of the seven rule-based agents. We paired each rule-based agent with a new instance of the same Rainbow DQN architecture described above, using the scores obtained in the paired games as training reward. The resulting agents will be collectively referred to as paired rainbow agents, and individually by a subscript indicating the name of the rule-based agent used during training e.g. $Rainbow_{Piers}$.

These agents can illustrate the performance that can be obtained when training a RL agent to play with a specific partner and, more interestingly, could help us explore whether some rule-based agents serve as better training partners than others, by comparing the performance of each of those agents with partners other than the ones they trained with (including themselves). These agents were trained for a total of 27.5 million steps each.

\subsection{Agent Paired with Unknown Partner for Training}

We also ran a set-up where the partner used for each game during training was sampled uniformly from the pool of seven rule-based agents. We call this agent $Rainbow_{All}$, as it had access, during training time, to all seven rule-based agents.

This agent can be used to illustrate an ad-hoc scenario where the agent has played with all the agents it will encounter during validation, but their identity is unknown both during training and validation. It was also trained for a total of 27.5 million steps.

\section{Results}

All results below were achieved by playing 1000 games between the relevant pair of agents after training. With the exception of self-play games involving the original rule-based agents in the Java environment for validation purposes, all games were played in the Python environment.

\subsection{Validation of Rule-Based Re-Implementations}

We validated our new rule-based agents by comparing the self-play score of each rule-based agent in the Java environment as implemented in~\cite{canaan2018towards} and the new versions implemented in Python for this paper. The results of this validation are shown on table~\ref{table:validation}.


\begin{table*}
\begin{center}
\begin{tabular}{ccccccccc}
& $IGGI$ & $Internal$ & $Outer$ & $Random$ & $VDB$ & $Flawed$ & $Piers$  & Average\\
$Rainbow_{SP1}$ & 4.03 & 3.80 & 4.28 & 0.00 & 5.98 & 0.03 & 7.71 & 3.69 \\
$Rainbow_{SP2}$ & 4.08 & 3.76 & 7.05 & 0.00 & 8.94 & 0.02 & 8.83 & 4.67 \\
$Rainbow_{SP3}$ & 3.64 & 2.44 & 5.99 & 0.00 & 8.00 & 0.01 & 7.74 & 3.97 \\
$Rainbow_{SP4}$ & 4.76 & 3.37 & 5.63 & 0.00 & 8.07 & 0.02 & 8.35 & 4.31\\
$Rainbow_{SP5}$ & 4.51 & 3.65 & 5.56 & 0.00 & 8.62 & 0.07 & 8.03 & 4.35\\
\end{tabular}
\end{center}
\caption{Results obtained when pairing each version of $Rainbow_{SP}$ agents with each rule-based agent. The maximum SD and SEM are 5.32 and 0.17 respectively.}
\label{table:SP with RB}
\end{table*}

Most agents are within 1 point of their original score on the Java environment. $VDB$ shows the greatest discrepancy with difference over 2 points. These results are, for the most part, comparable to the differences of 0.7 to 1.5 points by~\cite{walton2017evaluating} when they evaluated their own re-implementations of $Internal$, $Outer$ and $VDB$, including the fact that $VDB$ had the higher discrepancy (of 1.5) in that evaluation. Interestingly, we verified that the difference between $VDB_J$ and $VDB_P$ vanishes in the 3-player setting of the game, the one the agent was originally developed for t. 

Since no previous work has attempted to bridge the gap between the Java and Python environments, it is impossible to know how much of the discrepancy is due to errors on our side or due to possible subtle differences between the environments themselves. Regardless, most agents achieve comparable scores between the two versions (with the big exception of $VDB$ achieving better scores in our new Pyhton version) so they were deemed good enough for our purposes of providing RL agents with partners that implement simple, but reasonably effective (with the exception of $Random$ and $Flawed$) human-inspired strategies.

Note that $LegalRandom$ and $Flawed$ scoring zero is expected: both of these agents will too often play cards at random, almost certainly losing all lives and scoring zero as result. The average self-play score across all seven rule-based Python agents was 10.39, which is a useful number to compare to the performance achieved by rainbow agents paired with the rule-based agents: a lower number means that the rule-based agents would be better off, on average, paired with themselves.

To get a feel for whether these agents play well with each other, we also put the seven Python implementations of the rule-based agents to play among themselves. Table~\ref{table:rulebased matchups} shows the results of playing each match-up for 1000 games, randomizing which player goes in the starting position.

All agents except $Flawed$ and $LegalRandom$ achieve reasonable scores with other agents, with the agents that are better at self-play also performing better when paired with the others. Weaker agents seem to benefit from being paired with stronger agents, with $Internal$ and $Outer$ achieving better scores when paired with $IGGI$, $VDB$ and $Piers$ than their own self-play scores.

\subsection{Performance of Rainbow Agents trained through Self-Play}

We then took the five instances of $Rainbow$ trained through self-play and paired them with each other for evaluation. The purpose was to verify and quantify the observation made by Bard \textit{et al.}~\cite{bard2020hanabi} that independently trained versions of the agent play well with each other. The result of this evaluation can be seen on table~\ref{table:SelfPlay}, which confirms their observation.

All agents achieve similar performance with other agents as their self-play performances. Note that the first agent is, in fact, better off being paired with any of the other four agents than with itself, possibly due to having fewer training steps and the worst self-play performance. Agent 4 has both the best self-play performance and the best average score when paired with the other four agents, possibly doing to have the most training steps.

Next, we paired each of those five agents with the seven rule-based agents we implemented. The results can be seen on table~\ref{table:SP with RB}. Unfortunately, the Rainbow agents were unable to play well with the rule-based agents. While the rule-based agents have a combined self-play score of 10.39, the best $Rainbow_{SP}$ for playing with them was $Rainbow_{SP2}$, which scores 4.67 points on average. $Rainbow_{SP2}$ was the one with the most training time, but not the highest self-play score. While the added training time might have helped the agent face a greater diversity of scenarios, it is unclear whether added training time (on a self-play regime) would have a positive or negative effect in the long run: it is possible that, after some point, the agent would start learning strategies that improve self-play score at the cost of performance with other agents.

The best rule-based agents at playing with the $Rainbow_{SP}$ agents are $VDB$ and $Piers$, who score around 8 points each. These are also the only two agents in the pool (other than $Flawed$ and $LegalRandom$) that will play cards that aren't 100\% certain to be playable, which might fare comparatively well with the $Rainbow_{SP}$ agents' preference for hinting at playable cards.


\begin{table*}
\begin{center}
\begin{tabular}{ccccccccc}
& $IGGI$ & $Internal$ & $Outer$ & $Random$ & $VDB$ & $Flawed$ & $Piers$ & Average \\
$Rainbow_{IGGI}$ & 11.01 & 2.83 & 2.08 & 0 & 7.33 & 0 & 4.96 & 4.03\\
$Rainbow_{Internal}$ & 11.03 & 9.27 & 8.19 & 0 & 10.10 & 0.05 & 9.78 & 6.91\\
$Rainbow_{Outer}$ & 8.65 & 3.51 & 8.42 & 0 & 7.02 & 0.02 & 7.18 & 4.97 \\
$Rainbow_{Random}$ & 0 & 0 & 0 & 0 & 0.03 & 0 & 0.01  & 0.00\\
$Rainbow_{VDB}$ & 14.91 & 6.26 & 7.86 & 0 & 15.92 & 0 & 11.43 & 8.05  \\
$Rainbow_{Flawed}$ & 3.29 & 2.87 & 3.06 & 0.01 & 5.93 & 1.51 & 6.00 & 3.24 \\
$Rainbow_{Piers}$ & 14.37 & 4.57 & 8.43 & 0 & 10.03 & 0.007 & 15.21 & 7.52 \\
$Rainbow_{All}$ & 16.31 & 10.85 & 13.67 & 0.00 & 15.70 & 0.32 & 15.8 & 10.38\\
\end{tabular}
\end{center}
\caption{Scores obtained by matching each paired Rainbow agent (rows) with every rule-based agent (columns). Highest SD is 7.4 and highest SEM is 0.23. }
\label{table:paired mixed}
\end{table*}

\begin{table}
\begin{center}
\begin{tabular}{ccc}
\textbf{Rainbow Agent} & \textbf{Paired score} & \textbf{Original score}\\
$Rainbow_{IGGI}$ & 11.01 & 15.77 \\
$Rainbow_{Internal}$ & 9.27 & 10.01 \\
$Rainbow_{Outer}$ & 8.42 & 13.78 \\
$Rainbow_{Random}$ & 0 & 0.00 \\
$Rainbow_{VDB}$ & 15.92 & 16.12 \\
$Rainbow_{Flawed}$ & 1.51 & 0.00 \\
$Rainbow_{Piers}$ & 15.21 & 17.06 \\
\end{tabular}
\end{center}
\caption{Comparison of the score obtained by a paired rainbow agent when paired with its training partner (Paired Score) and the self-play score of its training partner (Original Score). Original score is the same as shown in table~\ref{table:validation} and shown here for comparison.  Highest SD is 5.18 and highest SEM is 0.16.}
\label{table:paired specific}
\end{table}

\subsection{Performance of Paired Agents}



Our final experiments concerned $Rainbow$ agents that were trained to play with one or more of the rule-based agents. These agents were expected to  play better with their respective partners than the ones trained on a self-play regime, but it was not clear whether they would also play better with unseen partners, and what their own self-play score would be.

Table~\ref{table:paired mixed} shows the results of pairing, after training, each $Rainbow$ agent that was trained with a rule-based partner with all rule-based partners. We also show the results for the $Rainbow_{All}$ agent, who played all rule-based agents uniformly during training.

The best agent for this scenario is, unsurprisingly, $Rainbow_{All}$, with an average of 9.78 across all seven pairings. Other than $Rainbow_{All}$, the two next best agents are the ones trained with $VDB$ and $Piers$, with average scores slightly above 7 points. Among the rule-based agents, $IGGI$ had the best pairings on average across the Rainbow agents, with 9.95 points, despite the fact that $Rainbow_{IGGI}$, who trained with it, came in only at third place among the Rainbow agents. A possible explanation is that $IGGI$ is completely deterministic and only acts on known (not inferred) information. This would make it a very stable partner during test time, but not the best partner for training, as playing with it wouldn't lead to many diverse situations.

Next, table~\ref{table:paired specific} shows the results when considering only the Rainbow agents when tested with their respective training partners, for comparison with the rule-based agents' original self-play score. The only Rainbow agent to achieve better performance with their partner than the original agent's self-play performance was $Rainbow_{Flawed}$. Some of the other agents are close to the original score and could conceivably achieve it with some extra training. 

Interestingly, the average of these seven pairings is only 8.76, lower than the 10.38 average of $Rainbow_{All}$. This is despite $Rainbow_{All}$ having only $1/7$ of the training time with each rule-based agent and not knowing, during test time, the identity of its partner. It is possible that being exposed to a variety of behaviors by all agents might be better (at least with the training budget given) than attempting to learn the ``perfect'' strategy for each partner.

\begin{table}
\begin{center}
\begin{tabular}{cc}
\textbf{Rainbow agent} & \textbf{Self-play score}\\
$Rainbow_{IGGI}$ & 0.37 \\
$Rainbow_{Internal}$ & 3.91 \\
$Rainbow_{Outer}$ & 0.37 \\
$Rainbow_{Random}$ & 0 \\
$Rainbow_{VDB}$ & 1.68 \\
$Rainbow_{Flawed}$ & 1.96 \\
$Rainbow_{Piers}$ & 4.17 \\
$Rainbow_{All}$ & 5.62 \\
\end{tabular}
\end{center}
\caption{Score of each paired Rainbow agent evaluated in self-play mode. Highest SD is 6.16 and highest SEM is 0.19.}
\label{table:paired self}
\end{table}

Finally, table~\ref{table:paired self} shows the score of each $Rainbow_{Paired}$ agent on self-play. All but one self-play scores are below 5, meaning most of these agents are worse at self-play than the $Rainbow_{SP}$ agents are at cooperating with the rule-based agents. $IGGI$ and $Outer$ are particularly poor training partners for this scenario (even worse than $Flawed$). These are the two most deterministic agents, who also never act on partial information. This might have the double effect of not only exposing the training agent to little variety of behavior, but also failing to expose it to situations where it is punished for acting in misleading ways. This could lead, for example, to an agent that learns to expect every hint received to be about a playable card (as both $IGGI$ and $Outer$ prefer doing) but who never learned not to give hints about unplayable cards (as neither $IGGI$ or $Outer$ will act -wrongly!- on this partial information). This is a recipe for disaster when the challenge is changed to self-play.


\section{Conclusion and future work}

We trained agents using the popular Rainbow DQN architecture in Hanabi using self-play, a single rule-based partner, and a mix of rule-based partners. Our results show that the agents trained with self-play fail to cooperate well with any of the rule-based agents. Agents trained to play with rule-based agents could, in some cases, obtain results comparable to the self-play scores of those rule-based agents, and could conceivably surpass that benchmark with more training. However, they also did not play well with agents not seen during training, wich includes themselves, as their self-play performance was very poor.

Hanabi owes its interest as an AI testbed in large part due to the challenge of ad-hoc play, where the unique nature of its hidden information and its restricted communication channel seem to require something resembling a theory of mind. However, our results, alongside previously published results on $ACHA$~\cite{bard2020hanabi} and $BAD$~\cite{foerster2018bayesian} indicate that currently published agents all fail to achieve reasonable scores with unseen partners, despite being able to achieve around 20 points or more (out of a maximum of 25) in the self-play setting. 

One course of action would be to inspect the games to investigate why the $Rainbow$ agents seems to converge on compatible policies when independently trained with self-play, but still fail spectacularly when paired with the rule-based agents. Is $Rainbow$  using conventions based on color? Is it learning some inflexible version of a convention such as ``only hint playable cards''? Is it more often misleading its partner into playing an illegal card or is it the one misinterpreting the other agent's hints?

Another question worth investigating is whether additional training time helps or hurts when playing with agents \emph{not} seen during training. At which point (if ever) does the agent start to overfit, improving performance when paired with the training agents at the cost of performance with heldout agents? Would  regularization techniques such as L1 and L2 regularization~\cite{ng2004feature} or dropout regularization~\cite{hinton2012improving} improve performance with unseen agents?

Yet another question is what is the impact of the choice of scoring scheme. An agent trained with the more lenient scoring scheme might perform better, even if evaluated on the strict scheme, as it would receive a more meaningful learning signal in the early stages of training, where scores will likely be all zero in the strict scheme. 

Finally, it would be interesting to see the impact of the selection of training partners. Rather than using a small set of hand-crafted agents, we could use as agents procedurally generated by a Quality Diversity algorithm such as is done in~\cite{canaan2019diverse}, tree-search agents such as~\cite{goodman2019re} or even other RL agents trained with diverse sets of hyperparameters and reward functions, such as the Starcraft League seen in~\cite{arulkumaran2019alphastar}. This would be especially interesting for training agents that, contrary to the architecture used in our paper, has access to a longer history of the game and is potentially able to identify and adapt to different partners based on their observed actions.


\section*{ACKNOWLEDGMENT}

Rodrigo Canaan gratefully acknowledges the financial support from Honda Research Institute Europe (HRI-EU). We also thank Chris DiMauro for maintaining the computers used for this research.

\bibliographystyle{IEEEtran}
\bibliography{bibfile}

\end{document}